\documentclass[journal,onecolumn,12pt]{IEEEtran}
\usepackage{amsmath,amsfonts}
\usepackage{array}
\usepackage[caption=false,font=normalsize,labelfont=sf,textfont=sf]{subfig}
\usepackage{textcomp}
\usepackage{stfloats}
\usepackage{url}
\usepackage{verbatim}
\usepackage{graphicx}
\usepackage{cite}
\usepackage{url}            
\usepackage{amsfonts}       
\usepackage{xcolor}         
\usepackage{wrapfig}
\usepackage{graphicx}
\usepackage{amsmath}
\usepackage[ruled]{algorithm2e}
\usepackage{bbm}
\usepackage{stmaryrd,scalerel}
\DeclareMathOperator*{\argmin}{argmin}
\newtheorem{definition}{Definition}

\usepackage{amssymb}
\newtheorem{theorem}{Theorem}
\newtheorem{assumption}{Assumption}
\newcommand{\rvphi}         {{\pmb{\phi}}}
\newcommand{\rvtau}         {{\pmb{\tau}}}
\makeatletter
\DeclareMathAlphabet\mathbfcal{OMS}{cmsy}{b}{n} 

\ifCLASSOPTIONcompsoc
  \usepackage[nocompress]{cite}
\else
  \usepackage{cite}
\fi
\ifCLASSINFOpdf
\else
\fi
\begin{document}
%
\title{Few-Shot Calibration of Set Predictors via \\Meta-Learned Cross-Validation-Based  \\Conformal Prediction}
%
%
%
%

\author{Sangwoo~Park, Kfir~M.~Cohen, Osvaldo~Simeone
\IEEEcompsocitemizethanks{\IEEEcompsocthanksitem Sangwoo Park, Kfir M. Cohen, and Osvaldo Simeone are with King's Communication, Learning, \& Information Processing (KCLIP) lab, Department of Engineering, King’s College London, London WC2R 2LS, U.K.\protect\\
E-mail: sangwoo.park@kcl.ac.uk \protect\\
Code is available at https://github.com/kclip/meta-XB.}
}
\IEEEtitleabstractindextext{%
\begin{abstract}
Conventional frequentist learning is known to yield poorly calibrated models that fail to reliably quantify the uncertainty of their decisions. Bayesian learning can improve calibration, but formal guarantees apply only under restrictive assumptions about correct model specification. Conformal prediction (CP) offers a general framework for the design of set predictors with calibration guarantees that hold regardless of the underlying data generation mechanism. However, when training data are limited, CP tends to produce large, and hence uninformative, predicted sets. This paper introduces a novel meta-learning solution that aims at reducing the set prediction size. Unlike prior work, the proposed meta-learning scheme, referred to as meta-XB, \emph{(i)} builds on cross-validation-based CP, rather than the less efficient validation-based CP; and \emph{(ii)}  preserves formal per-task calibration guarantees, rather than less stringent task-marginal guarantees. Finally, meta-XB is extended to adaptive non-conformal scores, which are shown empirically to further enhance marginal per-input calibration.
\end{abstract}

\begin{IEEEkeywords}
Conformal prediction, meta-learning, cross-validation-based conformal prediction, set prediction, calibration.
\end{IEEEkeywords}}

\maketitle

\IEEEdisplaynontitleabstractindextext

%
\IEEEpeerreviewmaketitle


%
%
%
%

\section{Introduction}\label{sec:intro}
\subsection{Context and Motivation}
\label{sec:context}
In modern application of artificial intelligence (AI), \emph{calibration} is often deemed as important as the standard criterion of (average) accuracy \cite{thelen2022comprehensive}. A well-calibrated model is one that can reliably quantify the uncertainty of its decisions \cite{guo2017calibration, hermans2021averting}. Information about uncertainty is critical when access to data is limited and AI decisions are to be acted on by human operators, machines, or other algorithms. Recent work on calibration for AI has focused on Bayesian learning, or related ensembling methods, as means to quantify epistemic uncertainty \cite{finn2018probabilistic, yoon2018bayesian, ravi2018amortized,jose2022information}. However, recent studies have shown the limitations of Bayesian learning when the assumed model likelihood or prior distribution are \emph{misspecified} \cite{masegosa2020learning}. Furthermore, exact Bayesian learning is computationally infeasible, calling for approximations such as Monte Carlo (MC) sampling \cite{robert1999monte} and variational inference (VI) \cite{blundell2015weight}. Overall, under practical conditions, Bayesian learning does not provide \emph{formal guarantees} of calibration.

\begin{figure}[t]
 \begin{center}\includegraphics[scale=0.22]{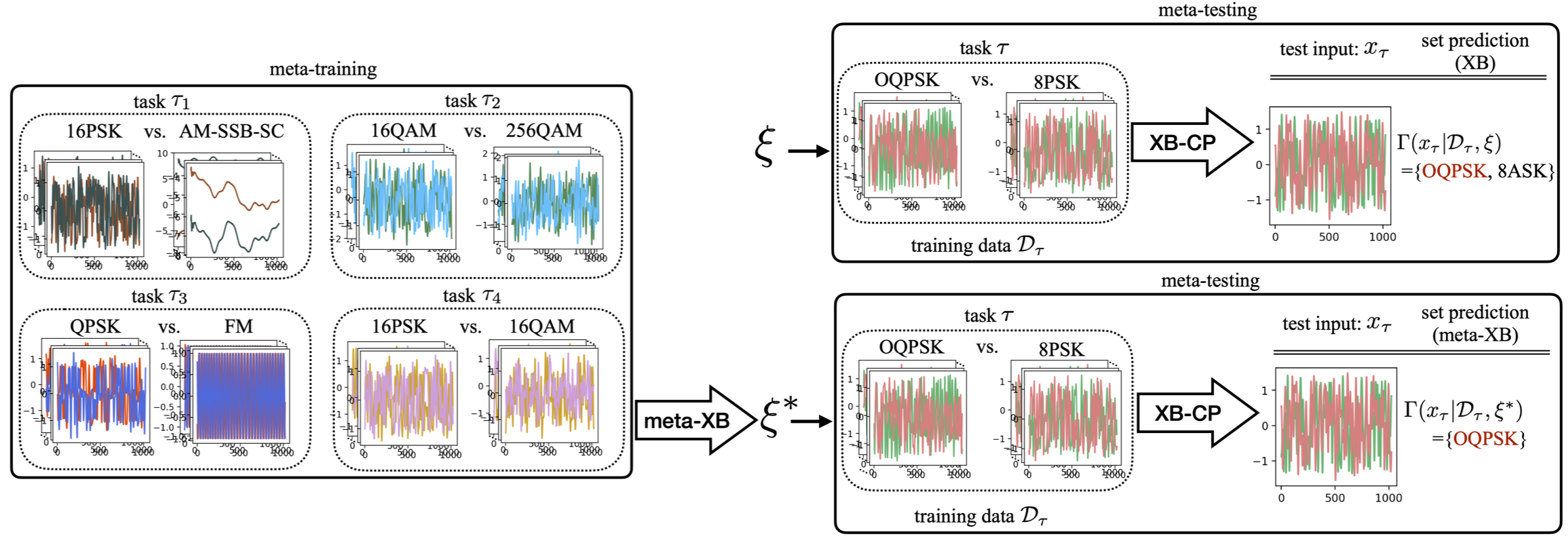} 
   \caption{Illustration of proposed meta-learned cross-validation-based CP (XB-CP) scheme, referred to as meta-XB. The example refers to the problem of classifying received radio signals depending on the modulation scheme used to generate it, e.g., QPSK or FM \cite{o2016convolutional, o2018over}. Based on data from multiple tasks, meta-XB optimizes a hyperparameter vector $\xi^*$ by minimizing the average set prediction size. As compared to conventional XB, shown on the top-right part of the figure, which uses a fixed hyperparameter vector $\xi$, meta-XB can achieve reduced set prediction size, while maintaining the per-task validity property \eqref{eq:per_task_marginal_validity}.} \label{fig:overall}
   \end{center}
 \vspace{-0.35cm}
\end{figure}

\emph{Conformal prediction (CP)} \cite{vovk2005algorithmic} provides a general framework for the calibration of (frequentist or Bayesian) probabilistic models. The formal calibration guarantees provided by CP hold irrespective of the (unknown) data distribution, as long as the available data samples and the test samples are exchangeable -- a weaker requirement than the standard i.i.d. assumption. As illustrated in Fig.~\ref{fig:overall}, CP produces \emph{set predictors}  that output a subset of the output space $\mathcal{Y}$ for each input $x$, with the property that the set contains the true output value with probability no smaller than a desired value $1-\alpha$ for $\alpha \in [0,1]$. 

Mathematically, for a given learning task $\tau$, assume that we are given a data set $\mathcal{D}_\tau$ with $N_\tau$ samples, i.e., $\mathcal{D}_\tau = \{z_\tau[i]\}_{i=1}^{N_\tau}$, where the $i$th sample $z_\tau[i]=(x_\tau[i],y_\tau[i])$ contains input $x_\tau[i] \in \mathcal{X}_\tau$ and target $y_\tau[i] \in \mathcal{Y}_\tau$.  
CP provides a \emph{set predictor} $\Gamma(\cdot|\mathcal{D}_\tau,\xi): \mathcal{X}_\tau \rightarrow 2^{\mathcal{Y}_\tau}$, specified by a hyperparameter vector $\xi$, that maps an input $x_\tau \in \mathcal{X}_\tau$ to a subset of the output domain $\mathcal{Y}_\tau$ based on a data set $\mathcal{D}_\tau$. Calibration amounts to the \emph{per-task validity} condition
\begin{align}
    \mathbb{P}(\mathbf{y}_\tau \in \Gamma(\mathbf{x}_\tau|\mathbfcal{D}_\tau,\xi)) \geq 1-\alpha,
    \label{eq:per_task_marginal_validity}
\end{align}
which indicates that the set predictor $\Gamma(\mathbf{x}_\tau|\mathbfcal{D}_\tau,\xi)$ contains the true target $\mathbf{y}_\tau$ with probability at least $1-\alpha$. In \eqref{eq:per_task_marginal_validity}, the probability $\mathbb{P}(\cdot)$ is taken over the ground-truth, exchangeable, joint distribution $p(\mathcal{D}_\tau,z_\tau)$, and bold letters represent random variables.

The most common form of CP, referred to as \emph{validation-based CP} (VB-CP), splits the data set into training and validation subsets \cite{vovk2005algorithmic}. The validation subset is used to calibrate the set prediction $\Gamma_\alpha^\text{VB}(x_\tau|\mathcal{D}_\tau,\xi)$ on a test example $x_\tau$ for a given desired miscoverage level $\alpha$ in \eqref{eq:per_task_marginal_validity}. The drawback of this approach is that validation data is not used for training, resulting in inefficient set predictors $\Gamma_\alpha^\text{VB}(x_\tau|\mathcal{D}_\tau,\xi)$ in the presence of a limited number $N_\tau$ of data samples. The average size of a set predictor $\Gamma(x_\tau|\mathcal{D}_\tau,\xi)$, referred to as \emph{inefficiency}, is defined as
\begin{align}
    \mathcal{L}_\tau(\xi) = \mathbb{E}\big| \Gamma(\mathbf{x}_\tau|\mathbfcal{D}_\tau,\xi) \big|, \label{eq: inefficiency}
\end{align}
where the average is taken with respect to the ground-truth joint distribution $p(\mathcal{D}_\tau,z_\tau)$. 


A more efficient CP set predictor was introduced by \cite{barber2021predictive} based on cross-validation. The \emph{cross-validation-based CP} (XB-CP) set predictor $\Gamma_\alpha^{K\text{-XB}}(x_\tau|\mathcal{D}_\tau,\xi)$ splits the data set $\mathcal{D}_\tau$ into $K$ folds to effectively use the available data for both training and calibration. XB-CP can also satisfy the per-task validity condition \eqref{eq:per_task_marginal_validity}\footnote{We refer here in particular to the jackknife-mm scheme presented in Section~2.2 of \cite{barber2021predictive}.}.

Further improvements in efficiency can be obtained via \emph{meta-learning} \cite{thrun1998lifelong}. Meta-learning jointly processes data from multiple learning tasks, say $\tau_1, \ldots, \tau_T$, which are assumed to be drawn i.i.d. from a task distribution $p(\tau)$. These data are used to optimize the hyperparameter $\xi$ of the set predictor $\Gamma(\mathbf{x}_\tau|\mathbfcal{D}_\tau,\xi)$ to be used on a new task $\tau \sim p(\tau)$. Specifically, reference \cite{fisch2021few} introduced a meta-learning-based method that modifies VB-CP. The resulting \emph{meta-VB} algorithm satisfies a looser validity condition with respect to the \emph{per-task} inequality \eqref{eq:per_task_marginal_validity}, in which the probability in \eqref{eq:per_task_marginal_validity} is no smaller than $1-\alpha$ only \emph{on average} with respect to the task distribution $p(\tau)$. 

\begin{figure}[t]
 \begin{center}\includegraphics[scale=0.18]{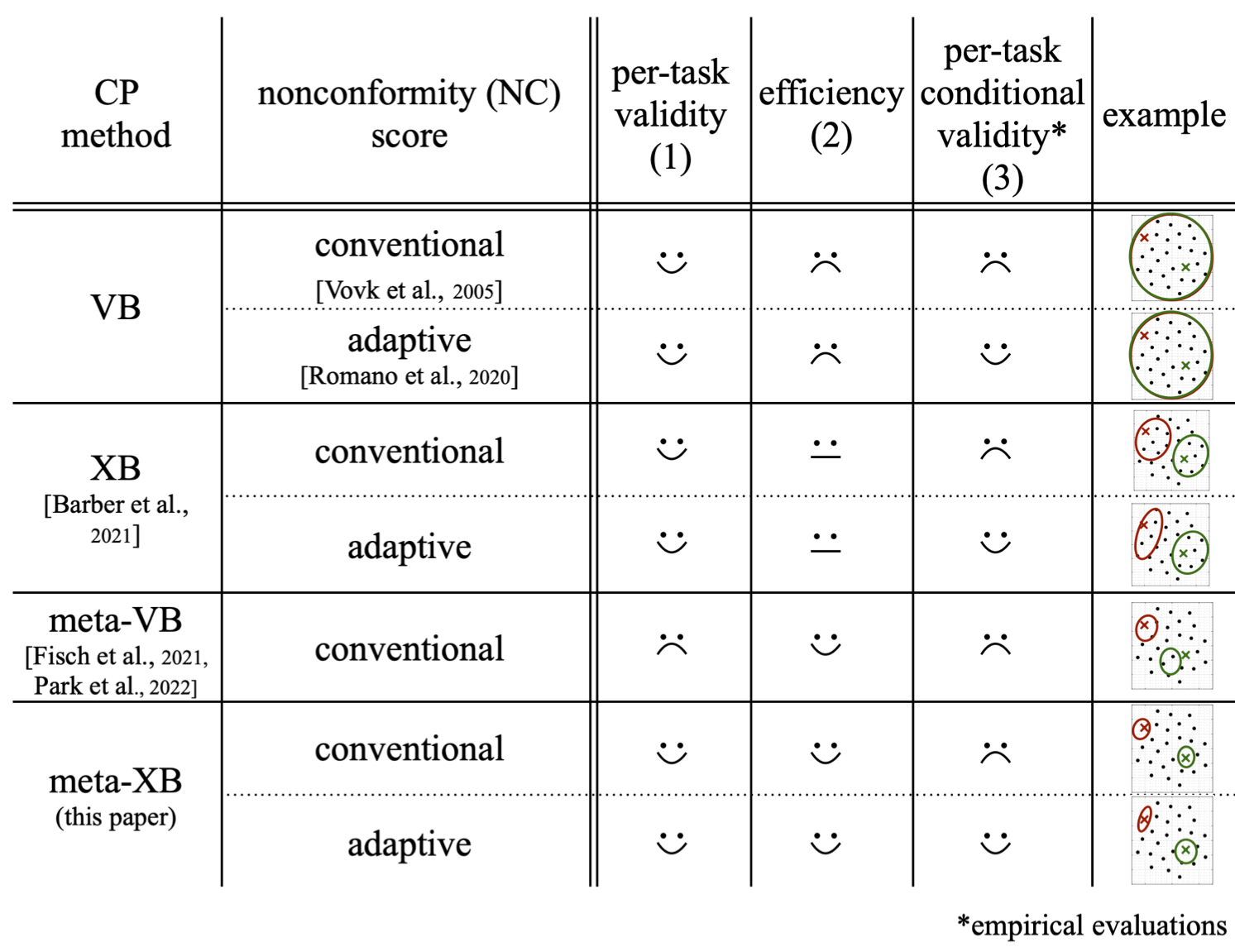} 
   \caption{Conformal prediction (CP)-based set predictors in the presence of limited data samples: Validation-based CP (VB-CP) \cite{vovk2005algorithmic} and the more efficient cross-validation-based CP (XB-CP) \cite{barber2021predictive} provide set predictors that satisfy the per-task validity condition (1); while previous works on meta-learning for VB-CP \cite{fisch2021few, park2022pac}, which aims at improving efficiency, do not offer validity guarantees when conditioning on a given task $\tau$. In contrast, the proposed meta-XB algorithm outputs efficient set predictors with guaranteed per-task validity. By incorporating adaptive NC scores \cite{romano2020classification}, meta-XB can also empirically improve per-input conditional validity (see \eqref{eq:per_task_cond_validity}). The last column illustrates efficiency, per-task validity, and per-task conditional validity for a simple example with possible outputs $y$ given by black dots, where the ground-truth outputs are given by the colored crosses and the corresponding set predictions by circles. Per-task validity (see (1)) holds if the set prediction includes the ground-truth output with high probability for each task $\tau$; while per-task conditional validity (see \eqref{eq:per_task_cond_validity}) holds when the set predictor is valid for each input. Conditional validity typically results in prediction sets of different sizes depending on the input \cite{romano2019conformalized, lin2021locally, leroy2021md, izbicki2020cd}. Inefficiency (see (2)) measures the average size of the prediction set.} \label{fig:cartoon}
   \end{center}
 \vspace{-0.35cm}
\end{figure}

\subsection{Main Contributions}
In this paper, we introduce a novel meta-learning approach, termed \emph{meta-XB}, with the aim of reducing the inefficiency \eqref{eq: inefficiency} of XB-CP, while preserving, unlike \cite{fisch2021few}, the per-task validity condition \eqref{eq:per_task_marginal_validity} for every task $\tau$. Furthermore, we incorporate in the design of meta-XB the \emph{adaptive nonconformity} (NC) scores introduced in \cite{romano2020classification}. As argued in \cite{romano2020classification} for conventional CP, adaptive NC scores are empirically known to improve the \emph{per-task conditional validity} condition 
\begin{align}
    \mathbb{P}(\mathbf{y}_\tau \in \Gamma(\mathbf{x}_\tau|\mathbfcal{D}_\tau,\xi) | \mathbf{x}_\tau = x_\tau ) \geq 1-\alpha.
    \label{eq:per_task_cond_validity}
\end{align}
This condition is significantly stronger than \eqref{eq:per_task_marginal_validity} as it holds for any test input $x_\tau$. A summary of the considered CP schemes can be found in Fig.~\ref{fig:cartoon}.

Overall, the contribution of this work can be summarized as follows:
\begin{itemize}
    \item We introduce meta-XB, a meta-learning algorithm for XB-CP, that can reduce the average prediction set size \eqref{eq: inefficiency} as compared to XB-CP, while satisfying the per-task validity condition \eqref{eq:per_task_marginal_validity}, unlike existing meta-learning algorithms for CP;
    \item We incorporate adaptive NC scores \cite{romano2020classification} in the design of meta-XB, demonstrating via experiments that adaptive NC scores can enhance conditional validity as defined by condition \eqref{eq:per_task_cond_validity}.
\end{itemize}

\section{Definitions and Preliminaries}
In this section, we describe necessary background material on CP \cite{vovk2005algorithmic, balasubramanian2014conformal}, VB-CP \cite{vovk2005algorithmic}, XB-CP \cite{barber2021predictive}, and adaptive NC scores \cite{romano2020classification}.

\subsection{Nonconformity (NC) Scores}
At a high level, given an input $x_\tau$ for some learning task $\tau$, CP outputs a prediction set $\Gamma(x_\tau|\mathcal{D}_\tau,\xi)$ that includes all outputs $y\in \mathcal{Y}_\tau$ such that the pair $(x_\tau, y)$ \emph{conforms} well with the examples in the available data set $\mathcal{D}_\tau = \{ z_\tau[i]=(x_\tau[i],y_\tau[i]) \}_{i=1}^{N_\tau}$. We recall from Section~1 that $\xi$ represents a vector of hyperparameter. The key underlying assumption is that data set $\mathcal{D}_\tau$ and test pair $z_\tau=(x_\tau, y_\tau)$ are realizations of \emph{exchangeable} random variables $\mathbfcal{D}_\tau$ and $\mathbf{z}_\tau$.

\begin{assumption}
    \label{assump:exchangeable}
    For any learning task $\tau$, data set $\mathbfcal{D}_\tau$ and a test data point $\mathbf{z}_\tau$ are exchangeable random variables, i.e., the joint distribution $p(\mathcal{D}_\tau,z_\tau) = p(z_\tau[1],\ldots,z_\tau[N_\tau],z_\tau)$ is invariant to any permutation of the variables $\{ \mathbf{z}_\tau[1],\ldots,\mathbf{z}_\tau[N_\tau],\mathbf{z}_\tau \}$. Mathematically, we have the equality $p(z_\tau[1],\ldots,z_\tau[N_\tau+1]) = p(z_\tau[\pi(1)],\ldots,z_\tau[\pi(N_\tau+1)])$ with $z_\tau=z_\tau[N_\tau+1]$, for any permutation operator $\pi(\cdot)$. Note that the standard assumption of i.i.d. random variables satisfies exchangeability.
\end{assumption}

CP measures conformity via \emph{NC scores}, which are generally functions of the hyperparameter vector $\xi$, and are defined as follows. 
    \begin{definition} \emph{(NC score)}
    \label{def:NC_score} For a given learning task $\tau$, given a data set $\tilde{\mathcal{D}}_\tau = \{ \tilde{z}_\tau[i]=(\tilde{x}_\tau[i],\tilde{y}_\tau[i]) \}_{i=1}^{\tilde{N}_\tau} \subseteq \mathcal{D}_\tau$ with $\tilde{N}_\tau \leq N_\tau$ samples, a nonconformity (NC) score is a function $\text{NC}(z|\tilde{\mathcal{D}_\tau},\xi)$ that maps the data set $\tilde{\mathcal{D}}_\tau$ and any input-output pair $z=(x,y)$ with $x \in \mathcal{X}_\tau$ and $y \in \mathcal{Y}_\tau$ to a real number while satisfying the permutation-invariance property $\text{NC}(z|\{\tilde{z}_\tau[1],\ldots,\tilde{z}_\tau[\tilde{N}]\},\xi)=\text{NC}(z|\{\tilde{z}_\tau[\pi(1)],\ldots,\tilde{z}_\tau[\pi(\tilde{N})]\},\xi)$ for any permutation operator $\pi(\cdot)$.
    \end{definition} 
A good NC score should express how poorly the point $(x_\tau,y)$ ``conforms'' to the data set $\tilde{\mathcal{D}}_\tau$. The most common way to obtain an NC score is via a parametric two-step approach. This involves a \emph{training algorithm} defined by a conditional distribution $p(\phi|\tilde{\mathcal{D}}_\tau,\xi)$, which describes the output $\rvphi$ of the algorithm as a function of training data set $\tilde{\mathcal{D}}_\tau \subseteq \mathcal{D}_\tau$ and hyperparameter vector $\xi$. This distribution may describe the output of a stochastic optimization algorithm, such as stochastic gradient descent (SGD), for frequentist learning, or of a Monte Carlo method for Bayesian learning \cite{guedj2019primer, angelino2016patterns, simeone2022machine}. The hyperparameter vector $\xi$ may determine, e.g., learning rate schedule or initialization.

        \begin{definition} \label{def:conven_two_step} \emph{(Conventional two-step NC score)} For a learning task $\tau$, let $\ell_\tau(z|\phi)$ represent the loss of a machine learning model parametrized by vector $\phi$ on an input-output pair $z=(x,y)$ with $x\in\mathcal{X}_\tau$ and $y\in\mathcal{Y}_\tau$. Given a training algorithm $p(\phi|\tilde{\mathcal{D}}_\tau,\xi)$ that is invariant to permutation of the training set $\tilde{\mathcal{D}}_\tau$, a conventional two-step NC score for input-output pair $z$ given data set $\tilde{\mathcal{D}}_\tau$ is defined as
    \begin{align}
    \label{eq:nc_general_definition}
     \text{NC}(z|\tilde{\mathcal{D}}_\tau,\xi) := \mathbb{E}_{{\rvphi} \sim p(\phi|\tilde{\mathcal{D}}_\tau,\xi)}\big[ \ell_\tau(z|\rvphi) \big].
        \end{align}    
    \end{definition}

    Due to the permutation-invariance of the training algorithm, it can be readily checked that \eqref{eq:nc_general_definition} is a valid NC score as per Definition~\ref{def:NC_score}.

\subsection{Validation-Based Conformal Prediction (VB-CP)}
VB-CP \cite{vovk2005algorithmic} divides the data set $\mathcal{D}_\tau$ into a training data set ${\mathcal{D}}_\tau^\text{tr}$ of $N_\tau^\text{tr}$ samples and a validation data set ${\mathcal{D}}_\tau^\text{val}$ of $N_\tau^\text{val}$ samples with ${N}_\tau^\text{tr}+{N}_\tau^\text{val}=N_\tau$. It uses the training data set $\mathcal{D}_\tau^\text{tr}$ to evaluate the NC scores $\text{NC}(z|\mathcal{D}_\tau^\text{tr},\xi)$, while the validation data set $\mathcal{D}_\tau^\text{val}$ is leveraged to construct the set predictor $\Gamma_\alpha^\text{VB}(x_\tau|\mathcal{D}_\tau,\xi)$ as detailed next.

Given an input $x_\tau$, the prediction set $\Gamma_\alpha^\text{VB}(x_\tau|\mathcal{D}_\tau,\xi)$ of VB-CP includes all output values $y\in\mathcal{Y}_\tau$ whose NC score $\text{NC}(z=(x_\tau,y)|\mathcal{D}_\tau^\text{tr},\xi)$ is smaller than (or equal to) a fraction (at least) $\lfloor\alpha (N_\tau^\text{val}+1)\rfloor/N_\tau^\text{val}$ of the NC scores $\{\text{NC}(z_\tau[i]|\mathcal{D}_\tau^\text{tr},\xi)\}_{i=1}^{N_\tau^\text{val}}$ for validation data points $z_\tau[i] \in \mathcal{D}_\tau^\text{val}$.

\begin{definition}
        The $(1-\alpha)$-empirical quantile $Q_{1-\alpha}(\{a[i]\}_{i=1}^M)$ of $M$ real numbers $a[1],\ldots,a[M]$, with $a[i]\in \mathbb{R}$, is defined as the  $\big\lceil (1-\alpha)(M+1) \big\rceil$th smallest value in the set $\{a[1],\ldots,a[M],\infty \}$.
\end{definition}

With this definition, the set predictor for VB-CP can be thus expressed as 
\begin{align}
    \label{eq:VB_set_predictor}
    &\Gamma_\alpha^\text{VB}(x_\tau|\mathcal{D}_\tau,\xi) = \Big\{ y \in \mathcal{Y}_\tau \hspace{-0.1cm}\: : \hspace{-0.1cm}\: \text{NC}(z|{\mathcal{D}}_\tau^\text{tr},\xi)  \:\leq\:  Q_{1-\alpha} \big(\big\{ \text{NC}({z}_\tau[i]|{\mathcal{D}}_\tau^\text{tr},\xi)\big\}_{i=1}^{{N}_\tau^\text{val}} \big) \text { with } z=(x_\tau,y) \Big\}.
\end{align}

Intuitively, by the exchangeability condition, the empirical ordering condition among the NC scores used to define set \eqref{eq:VB_set_predictor} ensures the validity condition (1) \cite{vovk2005algorithmic}.
\begin{theorem} \label{ther:VB-CP} \cite{vovk2005algorithmic} Under Assumption~\ref{assump:exchangeable}, for any miscoverage level $\alpha \in [1/(N^\text{val}_\tau+1),1)$, given any NC score as per Definition~\ref{def:NC_score}, the VB-CP set predictor \eqref{eq:VB_set_predictor} satisfies the validity condition (1).
\end{theorem}

\subsection{Cross-Validation-Based Conformal Prediction (XB-CP)}
\label{sec:XB-CP}
In VB-CP, the validation data set is only used to compute the empirical quantile in \eqref{eq:VB_set_predictor}, and is hence not leveraged by the training algorithm $p(\phi|\mathcal{D}_\tau^\text{tr},\xi)$. This generally causes the inefficiency (2) of VB-CP to be large if number of data points, $N_\tau$, is small. XB-CP addresses this problem via $K$-fold cross-validation \cite{barber2021predictive}.
$K$-fold cross-validation partitions the per-task data set $\mathcal{D}_\tau=\{ z_\tau[i] \}_{i=1}^{N_\tau}$ into $K \geq 2$ disjoint subsets ${\mathcal{D}}_{\tau,1},\ldots,{\mathcal{D}}_{\tau,K}$ such that the condition $\bigcup_{k=1}^K {\mathcal{D}}_{\tau,k} = \mathcal{D}_\tau$ is satisfied. We define the \emph{leave-one-out} data set ${\mathcal{D}}_{\tau,\neg k}= {\bigcup_{k'=1,k' \neq k}^K} {\mathcal{D}}_{\tau,k'}$ that excludes the subset ${\mathcal{D}}_{\tau,k}$. We also introduce a mapping function $k: \{1,\ldots,N_\tau\} \rightarrow \{1,\ldots,K\}$ to identify the subset ${\mathcal{D}}_{\tau,k(i)}$ that includes the sample $z_\tau[i]$, i.e., $z_\tau[i] \in {\mathcal{D}}_{\tau,k(i)}$. 

We focus here on a variant of XB-CP that is referred to as min-max jacknife+ in \cite{barber2021predictive}. This variant has stronger validity guarantees than the jacknife+ scheme also studied in \cite{barber2021predictive}. Accordingly, given a test input $x_\tau$, XB-CP computes the NC score for a candidate pair $z=(x_\tau,y)$ with $y \in \mathcal{Y}_\tau$ by taking the minimum NC score $\text{NC}(z|{\mathcal{D}}_{\tau,\neg k},\xi)$ over all possible subsets $k \in \{1,\ldots,K\}$, i.e., as $\min_{k \in \{1,\ldots,K\} } \text{NC}(z|\mathcal{D}_{\tau,\neg k},\xi)$. Furthermore, for each data point $z_\tau[i] \in \mathcal{D}_\tau$, the NC score is evaluated by excluding the subset ${\mathcal{D}}_{\tau,k(i)}$ as $\text{NC}(z_\tau[i]|\mathcal{D}_{\tau,\neg k(i)},\xi)$. Note that evaluating the resulting $N_\tau+1$ NC scores requires running the training algorithm $p(\phi| \mathcal{D}_{\tau,\neg k} ,\xi)$ $K$ times, once for each subset $\mathcal{D}_{\tau,\neg k}$. Finally, a candidate $y \in \mathcal{Y}_\tau$ is included in the prediction set if the NC score for $z=(x_\tau,y)$ is smaller (or equal) than for a fraction (at least) $\lfloor \alpha'(N_\tau+1)\rfloor/N_\tau$ of the validation data points with $\alpha' = \alpha - \frac{1-K/{N_\tau}}{K+1}$. 

Overall, given data set $\mathcal{D}_\tau=\{ z_\tau[i] =(x_\tau[i],y_\tau[i])\}_{i=1}^{N_\tau}$ and test input $x_\tau \in \mathcal{X}_\tau$, \emph{$K$-fold XB-CP} produces the set predictor
    \begin{align}
    \label{eq:cross_val_set_predictor_indicator}
        \Gamma_\alpha^{K\text{-XB}}(x_\tau|\mathcal{D}_\tau,\xi)= \Big\{ y \in \mathcal{Y}_\tau \: : \: \sum_{i=1}^{N_\tau} \mathbf{1} &\Big( \min_{k\in\{ 1,\ldots,K\}}\text{NC}(z|{\mathcal{D}}_{\tau,\neg k},\xi) \\&\leq  \text{NC}(z_\tau[i]|{\mathcal{D}}_{\tau,\neg k(i)},\xi)  \Big)\:\geq\: \lfloor \alpha'(N_\tau+1) \rfloor \text{ with } z=(x_\tau,y) \Big\},\nonumber
    \end{align}
    where $\mathbf{1}(\cdot)$ is the indicator function ($\mathbf{1}(\text{true})=1 $ and $\mathbf{1}(\text{false})=0$).

    \begin{theorem} 
    \label{ther:XB-CP} \cite{barber2021predictive} Under Assumption~\ref{assump:exchangeable}, for any miscoverage level $\alpha \in \big[\frac{1}{N_\tau+1} + \frac{1-K/N_\tau}{K+1},1\big)$, given any NC score as per Definition~\ref{def:NC_score}, the XB-CP set predictor \eqref{eq:cross_val_set_predictor_indicator} satisfies the validity condition (1).
\end{theorem}

While a proof of Theorem~\ref{ther:XB-CP} for $K=N_\tau$ can be found in \cite{barber2021predictive}, the general case for $K < N_\tau$ follows from the same proof techniques in \cite{barber2021predictive} and is included for completeness in Appendix~\ref{app:proof_XB_CP}.

\subsection{Adaptive Parametric NC Score}
\label{sec:adaptive_NC}

The CP methods reviewed so far achieve the per-task validity condition (1). In contrast, the per-input conditional validity \eqref{eq:per_task_cond_validity} is only attainable with 
strong additional assumptions on the joint distribution $p(\mathcal{D}_\tau,{x}_\tau)$ \cite{ vovk2012conditional, lei2014distribution}. 
However, the \emph{adaptive} NC score introduced by \cite{romano2020classification} is known to empirically improve the per-input conditional validity of VB-CP \eqref{eq:VB_set_predictor} and XB-CP \eqref{eq:cross_val_set_predictor_indicator}.

In this subsection, we assume that a model class of probabilistic predictors $p(y|x,\phi)$ is available, e.g., a neural network with a softmax activation in the last layer. To gain insight on the definition of adaptive NC scores, let us assume for the sake of argument that the ground-truth conditional distribution $p(y_\tau|x_\tau)$ is known. The most efficient (deterministic) set predictor satisfying the conditional coverage condition \eqref{eq:per_task_cond_validity} would then be obtained as the smallest-cardinality subset of target values in $\mathcal{Y}_\tau$ that satisfies the conditional coverage condition \eqref{eq:per_task_cond_validity}, i.e.,
        \begin{align}
            \Gamma^{*}_\alpha(x_\tau) = \argmin_{\Gamma \subseteq \mathcal{Y}_\tau} |\Gamma| \text{ s.t.} \sum_{y \in \Gamma} p(y|x_\tau) \geq 1-\alpha.
            \label{eq:optimal_genie_aided_set_predictor}
        \end{align}
Note that set \eqref{eq:optimal_genie_aided_set_predictor} can be obtained by adding values $y \in \mathcal{Y}_\tau$ to set predictor $\Gamma_\alpha^*(x_\tau)$ in order from largest to smallest value of $p(y|x_\tau)$ until the constraint in \eqref{eq:optimal_genie_aided_set_predictor} is satisfied.

In practice, the conditional distribution $p(y_\tau|x_\tau)$ is estimated via the model $p(y_\tau|x_\tau,\phi)$ where the parameter vector $\phi$ is produced by a training algorithm $p(\phi|\tilde{\mathcal{D}}_\tau,\xi)$ applied to some training data set $\tilde{\mathcal{D}}_\tau$. This yields the \emph{na\"ive} set predictor
        \begin{align}
        \label{eq:uncal_set_predictor}
        &\Gamma^{\text{na\"ive}}_{\alpha^\text{na\"ive}}(x_\tau|\tilde{\mathcal{D}}_\tau,\xi) = \argmin_{\Gamma \subseteq \mathcal{Y}_\tau} |\Gamma| \text{ s.t.} \sum_{y \in \Gamma} \mathbb{E}_{\rvphi \sim p(\phi|\tilde{\mathcal{D}}_\tau,\xi)}p(y|x_\tau,\rvphi) \geq 1-\alpha^\text{na\"ive},
        \end{align}
where we have used for generality the ensemble predictor obtained by averaging over the output $\rvphi \sim p(\phi|\tilde{\mathcal{D}}_\tau,\xi)$ of the training algorithm. Unless the likelihood model is perfectly calibrated, i.e., unless the equality $p(y_\tau|x_\tau)=\mathbb{E}_{\rvphi \sim p(\phi|\tilde{\mathcal{D}}_\tau,\xi)}[p(y_\tau|x_\tau,\phi)]$ holds, there is no guarantee that the set predictor in \eqref{eq:uncal_set_predictor} satisfies the conditional coverage condition \eqref{eq:per_task_cond_validity} or the marginal coverage condition (1) with $\alpha=\alpha^\text{na\"ive}$.

To tackle this problem, \cite{romano2020classification} proposed to apply VB-CP or XB-CP with a modified NC score inspired by the na\"ive prediction \eqref{eq:uncal_set_predictor}.

\begin{definition} \label{def:adaptive_NC} \emph{(Adaptive NC score)} For a learning task $\tau$, given a training algorithm $p(\phi|\tilde{\mathcal{D}}_\tau,\xi)$ that is invariant to permutation of the training set $\tilde{\mathcal{D}}_\tau$, the adaptive NC score for input-output pair $z=(x,y)$ with $x\in \mathcal{X}_\tau$ and $y\in\mathcal{Y}_\tau$ given data set $\tilde{\mathcal{D}}_\tau$, is defined as 
     \begin{align}
     \label{eq:nc_adaptive}
     \text{NC}^\text{ada}(z|\tilde{\mathcal{D}}_\tau,\xi) \hspace{-0.05cm}=\hspace{-0.2cm} \max_{\alpha^\text{na\"ive} \in [0,1]} \alpha^\text{na\"ive} \text{ s.t. } y \in \Gamma_{\alpha^\text{na\"ive}}^\text{na\"ive}(x_\tau|\tilde{\mathcal{D}}_\tau,\xi).
    \end{align}
    \end{definition}
    Intuitively, if the adaptive NC score is large, the pair $z$ does not conform well with the probabilistic model $\mathbb{E}_{\rvphi \sim p(\phi|\tilde{\mathcal{D}}_\tau,\xi) } p(y|x,\rvphi) $ obtained by training on set $\tilde{\mathcal{D}}_\tau$. The adaptive NC score satisfies the condition in Definition~\ref{def:NC_score}, and hence by Theorems~\ref{ther:VB-CP} and \ref{ther:XB-CP}, the set predictors \eqref{eq:VB_set_predictor} and \eqref{eq:cross_val_set_predictor_indicator} for VB-CP and XB-CP, respectively, are both valid when the adaptive NC score is used. Furthermore, \cite{romano2020classification} demonstrated improved conditional empirical coverage performance as compared to the conventional two-step NC score in Definition~\ref{def:conven_two_step}. This may be seen as a consequence of the conditional validity of the na\"ive predictor \eqref{eq:uncal_set_predictor} under the assumption of a well-calibrated model. 
    
    The adaptive NC score \eqref{eq:nc_adaptive} can be equivalently expressed as 
        \begin{align}
        \label{eq:nc_adaptive_implement}
            &\text{NC}^\text{ada}(z|\tilde{\mathcal{D}}_\tau,\xi) =\sum_{y'\in\mathcal{Y}_\tau} \mathbf{1} \big( p(y'|x,\tilde{\mathcal{D}}_\tau,\xi) \geq p(y|x,\tilde{\mathcal{D}}_\tau,\xi)  \big) p(y'|x,\tilde{\mathcal{D}}_\tau,\xi),
        \end{align}
        where we have used the notation $p(y|x,\tilde{\mathcal{D}}_\tau,\xi):=\mathbb{E}_{\rvphi \sim p(\phi|\tilde{\mathcal{D}}_\tau,\xi)}p(y|x,\rvphi)$. 



\section{Meta-Learning Algorithm for XB-CP (Meta-XB)}
In this section, we introduce the proposed meta-XB algorithm. We start by describing the meta-learning framework.

\subsection{Meta-Learning}
Up to now, we have focused on a single task $\tau$. Meta-learning utilizes data from multiple tasks to enhance the efficiency of the learning procedure for new tasks. Following the standard meta-learning formulation \cite{baxter2000model, amit2018meta}, as anticipated in Section~1, the learning environment is characterized by a task distribution $p(\tau)$ over the task identifier $\rvtau$. Given $T$ meta-training tasks realizations $\rvtau_1=\tau_1,\ldots,\rvtau_T=\tau_T$ drawn i.i.d. from the task distribution $p(\tau)$, the \emph{meta-training data set} $\mathcal{D}_{\tau_{1:T}}:=\{\{\mathcal{D}_t^j,{z}_t^j\}_{j=1}^{M_t}\}_{t=1}^T$ consists of $M_t$ realizations $\{\mathcal{D}_t^j,z_t^j\}_{j=1}^{M_t}$ of data sets $\mathcal{D}_{\tau_t}^j=\mathcal{D}_t^j$ with $N_{\tau_t}=N_t$ examples and test sample $z_{\tau_t}^j=z_t^j$ for each task $\tau_t$. Pairs $\{\mathcal{D}_t^j,{z}_t^j\}_{j=1}^{M_t}$ are generated i.i.d. from the joint distribution $p(\mathcal{D}_{\tau_t},z_{\tau_t})$, satisfying Assumption~\ref{assump:exchangeable} for all tasks $t$. 

The goal of meta-learning for CP is to optimize the vector of hyperparameter $\xi$ based on the meta-training data $\mathcal{D}_{\tau_{1:T}}$, so as to obtain a more efficient set predictor $\Gamma(x_\tau|\mathcal{D}_\tau,\xi)$. While reference \cite{fisch2021few} proposed a meta-learning solution for VB-CP \cite{vovk2005algorithmic}, here we introduce a meta-learning method for XB-CP.



\subsection{Meta-XB}
Meta-XB aims at finding a hyperparameter vector $\xi$ that minimizes the average size of the prediction set $\Gamma_\alpha^{K\text{-XB}}(x_\tau|\mathcal{D}_\tau,\xi)$ in \eqref{eq:cross_val_set_predictor_indicator} for tasks $\tau$ that follow the distribution $p(\tau)$. To this end, it addresses the problem of minimizing the empirical average of the sizes of the prediction sets $\Gamma_\alpha^{K\text{-XB}}(x_{\tau_t}|\mathcal{D}_{\tau_t},\xi)$ across the meta-training tasks $\tau_1,\ldots,\tau_T$ over the hyperparameter vector $\xi$. This amounts to the optimization
        \begin{align}
        \label{eq:meta_obj}
         \xi^* = \arg\min_{\xi} \frac{1}{T} \sum_{t=1}^T \frac{1}{M_t} \sum_{j=1}^{M_t} \big|  \Gamma_\alpha^{K\text{-XB}}(x_t^{j}|\mathcal{D}_t^j,\xi)  \big|,
    \end{align}
where the first sum is over the meta-training tasks and the second is over the available data for each task. 
By \eqref{eq:cross_val_set_predictor_indicator}, the size of the prediction set $|\Gamma_\alpha^{K\text{-XB}}(x|\mathcal{D},\xi)|$ is not a differentiable function of the hyperparameter vector $\xi$. Therefore, in order to address \eqref{eq:meta_obj} via gradient descent, we introduce a differentiable \emph{soft inefficiency} criterion by replacing the indicator function with the sigmoid $\sigma(u) = (1+\exp(-u/c_{\sigma}))^{-1}$ for some $c_\sigma > 0$; the quantile $Q_{1-\alpha}(\cdot)$ with a differentiable soft empirical quantile $\hat{Q}_{1-\alpha}(\cdot)$; and the minimum operator with the softmin function \cite{goodfellow2016deep}.

\begin{figure}[t]
        \centering
        \includegraphics[width=0.4\textwidth]{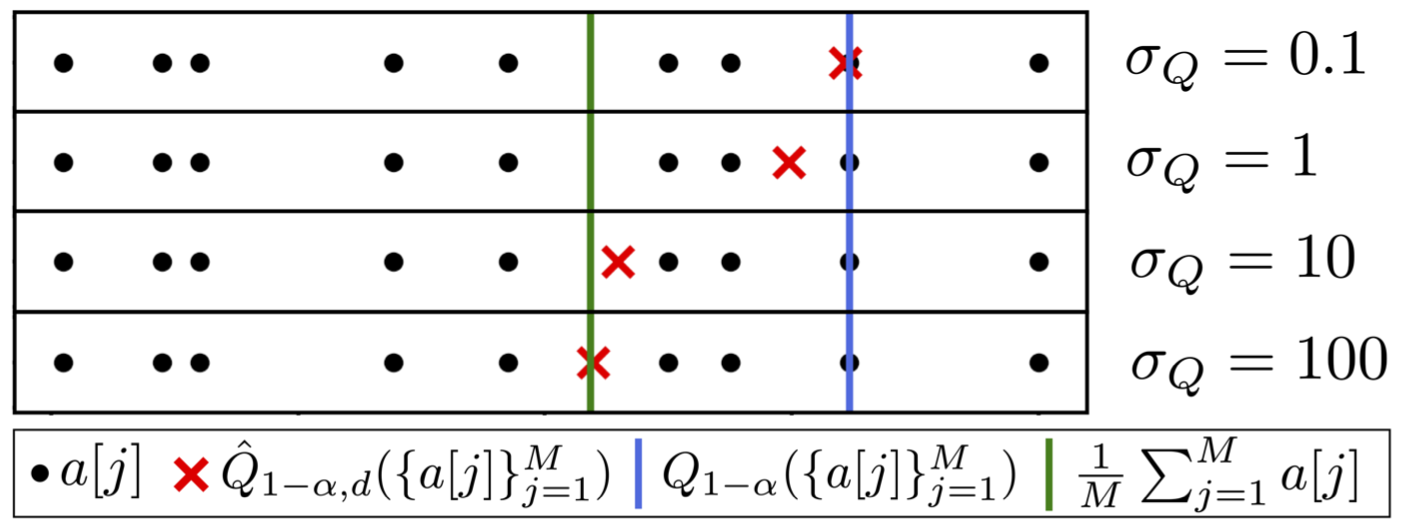}
        \caption{Trade-off between smoothness and accuracy of soft quantile. Dots represent the $M$ input values, the blue line is the true empirical quantile $Q_{1-\alpha}(\cdot)$, and the green line is the mean of the $M$ input values. }
        \label{fig:soft_quan_diagram}
        \end{figure}
For an input set $\{ a[j] \}_{j=1}^M$, the softmin function is defined as \cite[Section~6.2.2.3]{goodfellow2016deep} 
    \begin{align}
        \label{eq:softmin}
        &\text{softmin} \big(\{ a[j]\}_{j=1}^M\big) = \sum_{j=1}^M a[j]  \frac{\text{exp}(-a[j]/c_S)}{\sum_{i=1}^M \text{exp}(-a[i]/c_S)},
    \end{align} 
    for some $ c_S>0$. Finally, given an input set $\{ a[1],\ldots,a[M] \}$, the soft empirical quantile $\hat{Q}_{1-\alpha}(\cdot)$ is defined as
    \begin{align}
        \label{eq:softquan}
        &\hat{Q}_{1-\alpha}  \big(\{ a[j]\}_{j=1}^{M}\big) = \sum_{j=1}^{M+1} a[j]  \frac{\text{exp}(- \rho_{1-\alpha}(a[j]| \{ a[j] \}_{j=1}^{M+1}) /c_Q)}{\sum_{i=1}^{M+1} \text{exp}(- \rho_{1-\alpha}(a[i]| \{ a[j] \}_{j=1}^{M+1} ) /c_Q)},
    \end{align} 
for some $c_Q>0$ and $a[M+1] = \max(\{a[j]\}_{j=1}^M)+\delta$ for some $\delta>0$, where we have used the \emph{pinball loss} $\rho_{1-\alpha}(a| \{ a[1],\ldots,a[M] \} ) $ \cite{koenker1978regression}
\begin{align}
    \label{eq:pinball_loss}
    &\rho_{1-\alpha}(a| \{ a[j] \}_{j=1}^M ) = \alpha \sum_{j=1}^M \text{ReLU}(a - a[j]) + (1-\alpha) \sum_{j=1}^M \text{ReLU}(a[j]-a),
\end{align}
with $\text{ReLU}(a)=\max(0,a)$. With these definitions, the soft inefficiency metric is derived from \eqref{eq:cross_val_set_predictor_indicator} as follows (see details in Appendix~\ref{app:proof_soft_ineff}).

\begin{definition}
    \label{def:soft_ineff}
     Given a data set $\mathcal{D}_\tau$ and a test input $x_\tau$, the soft inefficiency for the $K$-fold XB-CP predictor \eqref{eq:cross_val_set_predictor_indicator} is defined as 
     \begin{align}
     \label{eq:soft_ineff_2}
         &|{\hat{\Gamma}}_\alpha^{K\text{-XB}}(x_\tau|\mathcal{D}_\tau,\xi)| \\&=  \sum_{y \in \mathcal{Y}} \sigma\Big(  \hat{Q}_{1-\alpha'}\Big(\Big\{ \text{NC}(z_\tau[i]|{\mathcal{D}}_{\tau,\neg k(i)},\xi) - \text{softmin}\big( \{ \text{NC}((x_\tau,y)|{\mathcal{D}}_{\tau,\neg k},\xi) \}_{k=1}^{K}\big)  \Big\}_{i=1}^N\Big) \Big),\nonumber
     \end{align} 
     where $\alpha' = \alpha - \frac{1-K/N_\tau}{K+1}$ and $c_\sigma,c_S,c_Q>0$. 
    \end{definition}
    The parameters $c_\sigma,c_S$, and $c_Q$ dictate the trade-off between smoothness and accuracy of the approximation $|{\hat{\Gamma}}_\alpha^{K\text{-XB}}(x_\tau|\mathcal{D}_\tau,\xi)|$ with respect to the true inefficiency $|{{\Gamma}}_\alpha^{K\text{-XB}}(x_\tau|\mathcal{D}_\tau,\xi)|$: As $c_\sigma,c_S,c_Q \rightarrow 0$, the approximation becomes increasingly accurate for any $\delta>0$, as long as we have $\alpha \in \big[\frac{1}{N_\tau+1} + \frac{1-K/N_\tau}{K+1},1\big)$, but the function $|{\hat{\Gamma}}_\alpha^{K\text{-XB}}(x_\tau|\mathcal{D}_\tau,\xi)|$ is increasingly less smooth (see Fig.~\ref{fig:soft_quan_diagram} for an illustration of the accuracy of the soft quantile).
    
    Replacing the soft inefficiency \eqref{eq:soft_ineff_2} into problem \eqref{eq:meta_obj} yields a differentiable program when conventional two-step NC scores (Definition~\ref{def:conven_two_step}) are used. We address the corresponding problem via stochastic gradient descent (SGD), whereby at each iteration a batch of tasks and examples per task are sampled. The overall meta-learning procedure is summarized in Algorithm~\ref{alg:meta_learning_KXB}.
    \subsection{Meta-XB with Adaptive NC Scores}
    Adaptive NC scores are not differentiable. Therefore, in order to enable the optimization of problem \eqref{eq:meta_obj} with the soft inefficiency \eqref{eq:soft_ineff_2}, we propose to replace the indicator function $\mathbf{1}{(\cdot)}$ in \eqref{eq:nc_adaptive_implement}  with the sigmoid function $\sigma(\cdot)$. We also have found that approximating the number of outputs $y' \in \mathcal{Y}_\tau$ that satisfy \eqref{eq:nc_adaptive_implement} rather than direct application of sigmoid function empirically improves per-input coverage performance. 
    This yields the \emph{soft adaptive NC score} $\hat{\text{NC}}^\text{ada}(z|\tilde{\mathcal{D}}_\tau,\xi)$, which is detailed in Appendix~\ref{app:soft_adaptive_NC}. With the soft adaptive NC score, meta-XB is then applied as in Algorithm~\ref{alg:meta_learning_KXB}.

\subsection{Per-Task Validity of Meta-XB}  As mentioned in Section~1, existing meta-learning schemes for CP cannot achieve the per-task validity condition in (1), requiring an additional marginalization over distribution $p(\tau)$ \cite{fisch2021few} or achieving looser validity guarantees formulated as probably approximately correct (PAC)-bounds \cite{park2022pac}. In contrast, meta-XB has the following property.
\begin{theorem} 
    \label{ther:meta-XB} Under Assumption~\ref{assump:exchangeable}, for any miscoverage level $\alpha \in \big[\frac{1}{N_\tau+1} + \frac{1-K/N_\tau}{K+1},1\big)$, given any NC score (Definition~\ref{def:NC_score}), the XB-CP set predictor \eqref{eq:cross_val_set_predictor_indicator} with $\xi = \xi^*$ in \eqref{eq:meta_obj} satisfies the validity condition (1).
\end{theorem} 
Theorem~\ref{ther:meta-XB} is a direct consequence of Theorem~\ref{ther:XB-CP}, since meta-XB maintains the permutation-invariance of the training algorithm $p(\phi|\tilde{\mathcal{D}}_\tau,\xi^*)$ as required by Definition~\ref{def:conven_two_step}.

\begin{algorithm}[h] 
\DontPrintSemicolon
\smallskip
\KwIn{meta-training set $\mathcal{D}_{1:T}= \{ \mathcal{D}_{t} \}_{t=1}^T$; number of examples $\{N_t\}_{t=1}^T$ to be used for set prediction; step size hyperparameter $\kappa$; approximation parameter $c_S$ for softmin, $c_Q$ for soft quantile, and $c_\sigma$ for sigmoid; minibatch size for tasks $\tilde{T}$ and minibatch size for realization pairs $\tilde{M}_t$}
\KwOut{meta-learned hyperparameter vector $\xi^*$}
\vspace{0.15cm}
\hrule
\vspace{0.15cm}
{\bf initialize} hyperparameter vector $\xi$ \\
\While{{\em convergence criterion not met}}{
choose $\tilde{T}$ tasks randomly from  set $\{1,\ldots,T\}$ and denote the corresponding task set as $\tilde{\mathcal{T}}$ \\
\For{{\em each sampled task $t \in \tilde{\mathcal{T}}$}}{\vspace{0.1cm}
randomly sample $\tilde{M_t}$ pairs from the data set ${\mathcal{D}}_{\tau_t}$, i.e., $\{\mathcal{D}_t^j, z_t^{j} \}_{j\in \tilde{\mathcal{J}}_t}$ denoting the corresponding index set as $\tilde{\mathcal{J}}_t$, and compute the soft inefficiency
\begin{align}
    \label{eq:SGD_ineff}
    \hat{\mathcal{L}}_t(\xi) = \frac{1}{\tilde{M}_t}\sum_{j\in\tilde{\mathcal{J}}_t}|\hat{\Gamma}_\alpha^{K\text{-XB}}(x_t^{j}|\mathcal{D}_t^j,\xi)|.
\end{align}
}
update hyperparameter vector $\xi \leftarrow \xi - \kappa \sum_{t \in \tilde{\mathcal{T}}}\nabla_{\xi}\hat{\mathcal{L}}_t(\xi)$
}
return the optimized hyperparameter vector $\xi$
\caption{Meta-XB}
\label{alg:meta_learning_KXB}
\end{algorithm}

\section{Related Work}   
    \textbf{Bayesian learning and model misspecification.} 
    When the model is misspecified, i.e., when the assumed model likelihood or prior distribution cannot express the ground-truth data generating distribution \cite{masegosa2020learning}, Bayesian learning may yield poor generalization performance \cite{masegosa2020learning, morningstar2022pacm, wenzel2020good}. Downweighting the prior distribution and/or the likelihood, as done in generalized Bayesian learning \cite{knoblauch2019generalized, simeone2022machine} or in ``cold'' posteriors \cite{wenzel2020good}, improve the generalization performance. In order to mitigate the model likelihood misspecification, alternative variational free energy metrics were introduced by \cite{masegosa2020learning} via second-order PAC-Bayes bounds, and by \cite{morningstar2022pacm} via multi-sample PAC-Bayes bounds. Misspecification of the prior distribution can be also addressed via Bayesian meta-learning, which optimizes the prior from data in a manner similar to empirical Bayes \cite{mackay2003information}.
    
    \textbf{Bayesian meta-learning} While frequentist meta-learning has shown remarkable success in few-shot learning tasks in terms of accuracy \cite{finn2017model, snell2017prototypical}, improvements in terms of calibration can be obtained by Bayesian meta-learning that optimizes over a hyper-posterior distribution from multiple tasks \cite{amit2018meta, finn2018probabilistic, yoon2018bayesian, ravi2018amortized,nguyen2020uncertainty,jose2022information}. The hyper-prior can also be modelled as a stochastic process to avoid the bias caused by parametric models \cite{rothfuss2021meta}.



    \textbf{CP-aware loss.}  \cite{stutz2021learning} and \cite{einbinder2022training} proposed CP-aware loss functions to enhance the efficiency or per-input validity \eqref{eq:per_task_cond_validity} of VB-CP. The drawback of these solutions is that they require a large amount of data samples, i.e., $N_\tau \gg 1$, unlike the meta-learning methods studied here.

    \textbf{Per-input validity and local validity.} As discussed in Section~\ref{sec:adaptive_NC}, the per-input validity condition \eqref{eq:per_task_cond_validity} cannot be satisfied without strong assumptions on the joint distribution $p(\mathcal{D}_\tau,z_\tau)$ \cite{vovk2012conditional, lei2014distribution}. Given the importance of adapting the prediction set size to the input to capture heteroscedasticity \cite{romano2019conformalized, izbicki2020cd}, a looser \textbf{local validity} condition, which conditions on a subset of the input data space $A_{x_\tau} \subset \mathcal{X}_\tau$ containing the input $x_\tau$ of interest, i.e., $x_\tau \in A_{x_\tau}$, has been considered in \cite{lei2014distribution, foygel2021limits}. Choosing a proper subset $A_{x_\tau}$ becomes problematic especially in high-dimensional input space \cite{izbicki2020cd, leroy2021md}, and \cite{tibshirani2019conformal, lin2021locally} proposed to reweight the samples outside the subset $A_{x_\tau}$ by treating the problem as \emph{distribution-shift} between the data set $\mathcal{D}_\tau$ and the test input $x_\tau$. 

\begin{figure}[t]
 \begin{center}\includegraphics[scale=0.24]{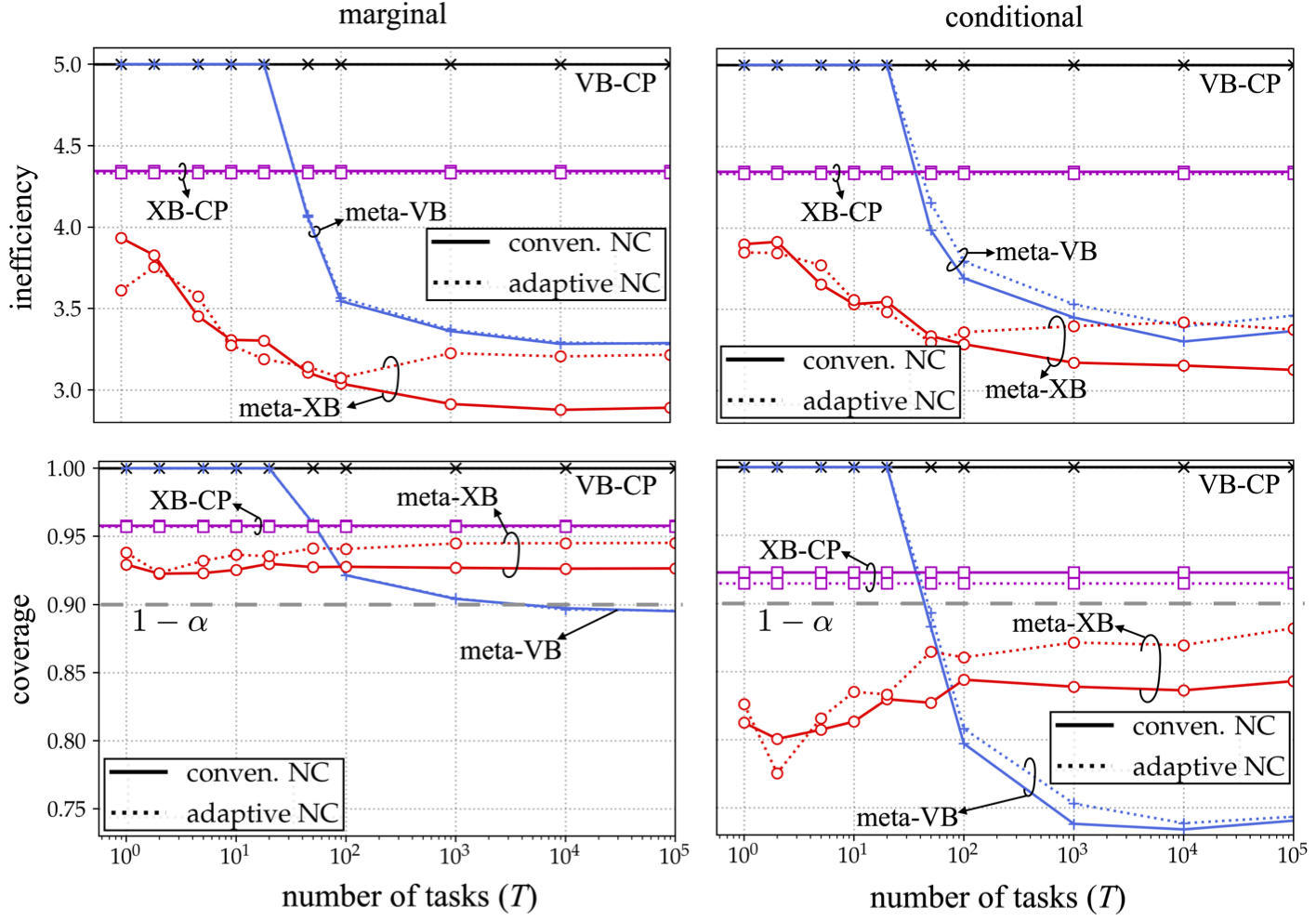} 
   \caption{Per-task inefficiency and coverage (left) and per-task conditional inefficiency and coverage (right) for VB-CP, XB-CP, meta-VB, and meta-XB for the synthetic-data example in \cite{romano2020classification}. Each data set $\mathcal{D}_\tau$ contains $N_\tau=9$ examples, while the meta-training data set $\mathcal{D}_{\tau_{1:T}}$ consists of $500$ examples per task from $M_t=50$ realizations.} \label{fig:toy_per_task}
   \end{center}
 \vspace{-0.35cm}
\end{figure}
\section{Experiments}
In this section, we provide experimental results to validate the performance of meta-XB in terms of \emph{(i)} per-task coverage $\mathbb{P}(\mathbf{y}_\tau \in \Gamma(\mathbf{x}_\tau|\mathbfcal{D}_\tau,\xi) )$; \emph{(ii)} per-task inefficiency (2); \emph{(iii)} per-task conditional coverage $\mathbb{P}(\mathbf{y}_\tau \in \Gamma(\mathbf{x}_\tau|\mathbfcal{D}_\tau,\xi) |\mathbf{x}_\tau = x_\tau )$; and \emph{(iv)} per-task conditional inefficiency $\mathbb{E}[ |\Gamma(\mathbf{x}_\tau|\mathbfcal{D}_\tau,\xi) ||\mathbf{x}_\tau = x_\tau ]$. To evaluate input-conditional quantities, we follow the approach in \cite[Section S1.2]{romano2020classification}. As benchmark schemes, we consider \emph{(i)} VB-CP, \emph{(ii)} XB-CP, and \emph{(iii)} meta-VB \cite{fisch2021few}, with either the conventional NC score (Definition~\ref{def:conven_two_step} with log-loss $\ell_\tau(z|\phi)$) or adaptive NC score  with Definition~\ref{def:adaptive_NC}). Note that meta-VB was described in \cite{fisch2021few} only for the conventional NC score, but the application of the adaptive NC score is direct. For all the experiments, unless specified otherwise, we consider a number of examples $N_\tau=9$ for the data set $\mathcal{D}_\tau$ and the desired miscoverage level $\alpha=0.1$. For the cross-validation-based set predictors XB-CP and meta-XB, we set number of folds to $K=N_\tau$. The aforementioned performance measures are estimated by averaging over $1000$ realizations of data set $\mathcal{D}_\tau$ and over $500$ realizations for the test sample $z_\tau$ of each task $\tau$. We report in this section the $100$ different per-task quantities which are computed from $100$ different tasks. 
During meta-training, for $T$ different tasks, we assume availability of $M_t(N_\tau+1)$ i.i.d. examples, from which we sample $\tilde{M}_t$ pairs $\{\mathcal{D}_t^j, z_t^{j} \}_{j\in \tilde{\mathcal{J}}_t}$ when computing inefficiency \eqref{eq:SGD_ineff}, with which we use Adam optimizer \cite{kingma2014adam} to update the hyperparameter vector $\xi$ via SGD. Lastly, we set the value of the approximation parameters $c_\sigma, c_S,$ and $c_Q$ to be one.


Following \cite{romano2020classification}, for VB-CP and XB-CP, we adopt a support vector classifier as training algorithm $p(\phi|\tilde{\mathcal{D}}_\tau,\xi)$ as it does not require any tuning of the hyperparameter vector $\xi$. In contrast, for meta-VB and meta-XB, we adopt a neural network classifier \cite{romano2019conformalized}, and set the training algorithm $p(\phi|\tilde{\mathcal{D}}_\tau,\xi)$ to output the last iterate of a pre-defined number of steps of GD ($1$, unless specified otherwise) with initialization given by the hyperparameter vector $\xi$ \cite{finn2017model}. Note that using full-batch GD ensures the permutation-invariance of the training algorithm as required by Definition~\ref{def:conven_two_step}.

All the experiments are implemented by PyTorch \cite{paszke2019pytorch} and ran over a GPU server with single NVIDIA A100 card.

\begin{figure}[t]
 \begin{center}\includegraphics[scale=0.28]{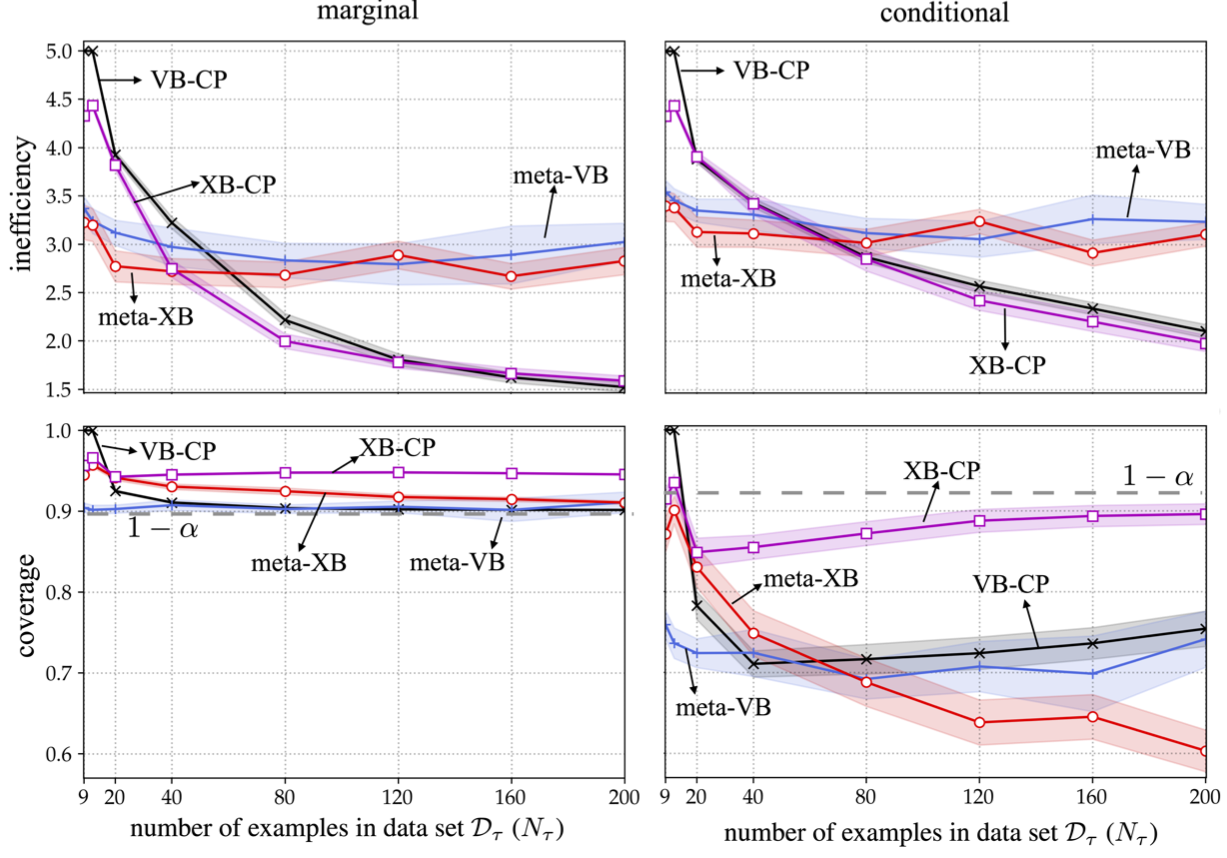} 
   \caption{Per-task inefficiency and coverage (left) and per-task conditional inefficiency and coverage (right) for VB-CP, XB-CP, meta-VB, and meta-XB for the synthetic-data example in \cite{romano2020classification} using adaptive NC scores. Different numbers of examples $N_\tau$ for each data set $\mathcal{D}_\tau$ is considered here,  while $T=1000$ tasks are used to generate the meta-training data set $\mathcal{D}_{\tau_{1:T}}$  consists of $(N_\tau+1)M_t$ examples per task from $M_t=50$ realizations. The shaded areas correspond to confidence intervals covering $95\%$ of the realized values.} \label{fig:toy_per_N_ori}
   \end{center}
 \vspace{-0.35cm}
\end{figure}

\subsection{Multinomial Model and Inhomogeneous Features}
We start with the synthetic-data experiment introduced in \cite{romano2020classification} in which the input $x \in \mathbb{R}^{10}$ is such that the first element equals $x_1 = 1$ with probability $1/5$ and $x_1 = -8$ otherwise, while the other elements $x_2,\ldots,x_{10}$ are i.i.d. standard Gaussian variables. For each task $\tau$, matrix $\tau \in \mathbb{R}^{10\times |\mathcal{Y}_\tau|}$ is sampled with i.i.d. standard Gaussian entries and the ground-truth conditional distribution $p(y_\tau|x_\tau)$ is defined as the categorical distribution
\begin{align}
    p(y_\tau=y|x_\tau) = \frac{\text{exp}(x_\tau^\top \tau_{y})}{\sum_{y'=1}^{|\mathcal{Y}_\tau|} \text{exp}(x_\tau^\top \tau_{y'})},
\end{align}
for $y \in \{1,\ldots,|\mathcal{Y}_\tau|\}$, where $\tau_{y} \in \mathbb{R}^{|\mathcal{Y}_\tau|}$ is the $y$th column of the task information matrix $\tau$. The number of classes is $|\mathcal{Y_\tau}|=5$ and neural network classifier consists of two hidden layers with Exponential Linear Unit (ELU) activation \cite{clevert2015fast} in the hidden
layers and a softmax activation in the last layer.


In Fig.~\ref{fig:toy_per_task}, we demonstrate the performance of the considered set predictors as a function of number of tasks $T$. Both meta-VB and meta-XB achieve lower inefficiency (2) as compared to the conventional set predictors VB-CP and XB-CP, as soon as the number of meta-training tasks is sufficiently large to ensure successful generalization across tasks \cite{yin2019meta, jose2020informationtheoretic}. For example, meta-XB with $T=100$ tasks obtain an average prediction set size of $3$, while XB-CP has an inefficiency larger than $4$. Furthermore, all schemes satisfy the validity condition (1), except for meta-VB for $T \gtrsim 10^4$, confirming the analytical results. Adaptive NC scores are seen to be instrumental in improving the conditional validity \eqref{eq:per_task_cond_validity} when used with meta-XB, although this comes at the cost of a larger inefficiency.

Next, we investigate the impact of number of per-task examples $N_\tau$ in data set $\mathcal{D}_\tau$ using adaptive NC scores. 
As shown in Fig.~\ref{fig:toy_per_N_ori}, the average size of the set predictors decreases as $N_\tau$ grows larger. In the few-examples regime, i.e., with $N_\tau \leq 40$, the meta-learned set predictors meta-VB and meta-XB outperform the conventional set predictors VB-CP and XB-CP in terms of inefficiency. However, when $N_\tau$ is large enough, i.e., when $N_\tau \geq 80$, conventional set predictors are preferable, as transfer of knowledege across tasks becomes unnecessary, and possibly deleterious \cite{amit2018meta} (see also \cite{park2020learning} for related discussions). In terms of conditional coverage, Fig.~\ref{fig:toy_per_N_ori} shows that cross-validation-based CP methods are preferable as compared to validation-based CP approaches.



\begin{figure}[t]
        \centering
        \includegraphics[width=0.34\textwidth]{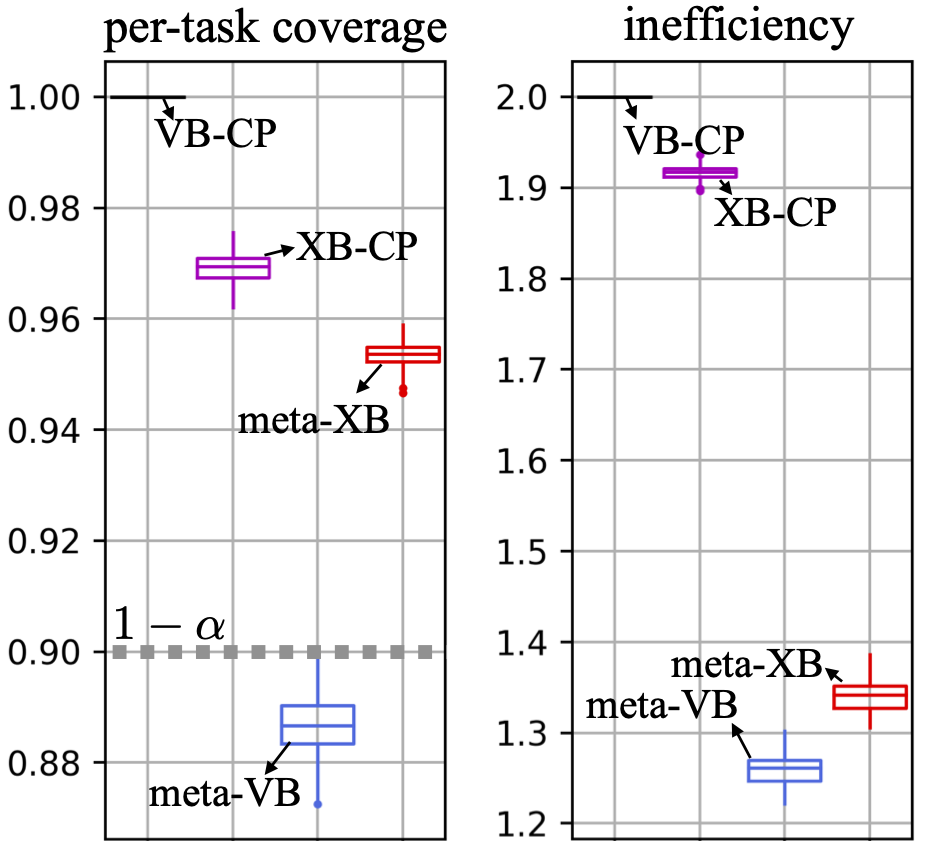}
        \caption{Per-task coverage and inefficiency of VB-CP, XB-CP, meta-VB, and meta-XB for modulation classification \cite{o2018over}. We consider $N_\tau=9$ examples for each data set $\mathcal{D}_\tau$. The boxes represent the $25\%$ (lower edge), $50\%$ (line within the box), and $75\%$ (upper edge) percentiles of the per-task performance metrics evaluated over $100$ different meta-test tasks.}
        \label{fig:ota}
        \end{figure}
\subsection{Modulation Classification}
We now consider the real-world \emph{modulation classification} example illustrated in Fig.~1, in which the goal is classifying received radio signals depending on the modulation scheme used to generate it \cite{o2016convolutional, o2018over}. 
The RadioML 2018.01A data set consists $98,304$ inputs with dimension $2\times 1024$, accounting for complex baseband signals sampled over $1024$ time instants, generated from $24$ different modulation types \cite{o2018over}. Each task $\tau$ amounts to the binary classification of signals from two randomly selected modulation types. Specifically, we divide the $24$ modulations types into $16$ classes used to generate meta-training tasks, and $8$ classes used to produce meta-testing tasks, following the standard data generation approach in few-shot classifications \cite{lake2011one, ravi2016optimization}. We adopt VGG16 \cite{simonyan2014very} as the neural network classifier as in \cite{o2018over}. Furthermore, for meta-VB and meta-XB, we apply a single GD step during meta-training and five GD steps during meta-testing \cite{finn2017model, ravi2018amortized}.

Fig.~\ref{fig:ota} shows per-task coverage and inefficiency for all schemes assuming conventional NC scores. While the conventional set predictors VB-CP and XB-CP produce large, uninformative set predictors that encompass the entire target data space $\mathcal{Y}_\tau$ of dimension $|\mathcal{Y}_\tau|=2$, the meta-learned set predictors meta-VB and meta-XB can significantly improve the prediction efficiency. However, meta-VB fails to achieve per-task validity condition (1), while the proposed meta-XB is valid as proved by Theorem~\ref{ther:meta-XB}.

\begin{figure}[t]
        \centering
        \includegraphics[width=0.34\textwidth]{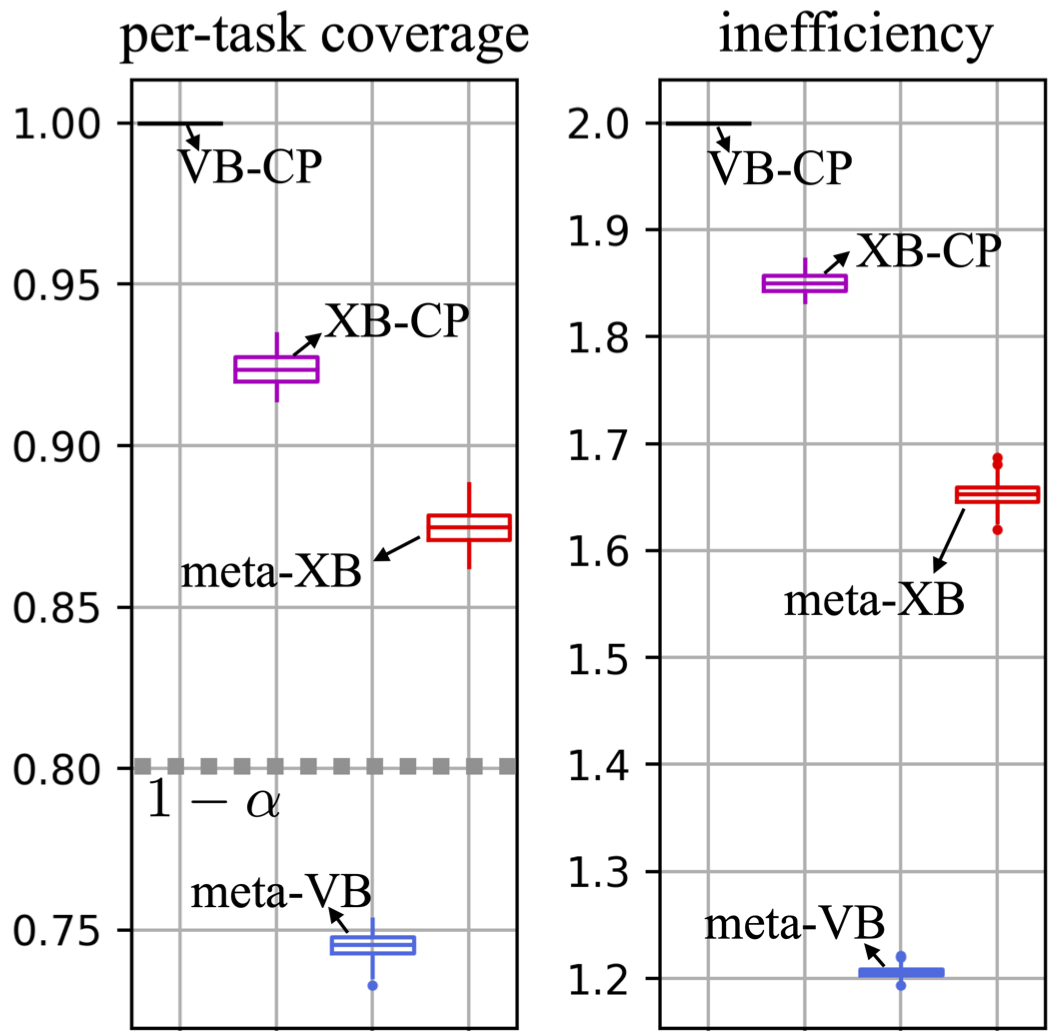}
        \caption{Per-task coverage and inefficiency of VB-CP, XB-CP, meta-VB, and meta-XB for \textit{mini}Imagenet \cite{vinyals2016matching}. We consider $N_\tau=4$ examples for each data set $\mathcal{D}_\tau$. The boxes represent the $25\%$ (lower edge), $50\%$ (line within the box), and $75\%$ (upper edge) percentiles of the per-task performance metrics evaluated over $100$ different meta-test tasks.} 
        \label{fig:mini}
        \end{figure}
\subsection{Image Classification}
Lastly, we consider image classification problem with the \textit{mini}Imagenet dataset \cite{vinyals2016matching} considering $N_\tau=4$ data points per task with desired miscoverage level $\alpha=0.2$. We consider binary classification with tasks being defined by randomly selecting two classes of images, and drawing training data sets by choosing among all examples belonging to the two chosen classes. Conventional NC scores are used, and the neural network classifier consists of the convolutional neural network (CNN) used in \cite{finn2017model}. For meta-VB and meta-XB, a single step GD update is used during meta-training, while five GD update steps are applied during meta-testing. Fig.~\ref{fig:mini} shows that meta-learning-based set predictors outperform conventional schemes. Furthermore, meta-VB fails to meet per-task coverage in contrast to the proposed meta-XB.

\section{Conclusion}
This paper has introduced meta-XB, a meta-learning solution for cross-validation-based conformal prediction that aims at reducing the average prediction set size, while formally guaranteeing per-task calibration. The approach is based on the use of soft quantiles, and it integrates adaptive nonconformity scores for improved input-conditional calibration. Through experimental results, including for modulation classification \cite{o2016convolutional, o2018over}, meta-XB was shown to outperform both conventional conformal prediction-based solutions and meta-learning conformal prediction schemes. Future work may integrate meta-learning with CP-aware training criteria \cite{stutz2021learning, einbinder2022training}, or with stochastic set predictors.
\section*{Acknowledgments}
The work of S. Park, K. M. Cohen, and O. Simeone was supported by the European Research Council (ERC) under the European Union’s Horizon 2020 Research and Innovation Programme (Grant Agreement No. 725731).
\appendices

\section{Proofs}
\subsection{Proof of Theorem~\ref{ther:XB-CP}}
\label{app:proof_XB_CP}

The proof mainly follows \cite[Section B.1]{barber2021predictive} and \cite[Section B.2.2]{barber2021predictive} with the following changes.

We first extend the result for regression problems in \cite{barber2021predictive} to classification, starting with the case $K=N_\tau$, for which the mapping function $k(\cdot)$ is the identity $k(i)=i$. Unlike \cite{barber2021predictive}, which defined ``comparison matrix'' of residuals for the regression problem, we consider a more general comparison matrix defined in terms of NC scores that can be applied for both classification and regression problems in a manner similar to \cite{romano2020classification}. Accordingly, we define the comparison matrix $A \in \{0,1\}^{(N_\tau +1)\times (N_\tau + 1) }$
\begin{align}
        A(i,j)\hspace{-0.1cm} = \hspace{-0.1cm}  \left.
      \begin{cases}
        &\hspace{-0.3cm}\mathbf{1}\Big( \min_{k \in \{1,\ldots,K+1\} } \text{NC}(z_\tau[i]| {\mathcal{D}}_{\tau, \neg (k(i), k)},\xi  )  > \text{NC}(z_\tau[j] | {\mathcal{D}}_{\tau, \neg (k(i), k(j))},\xi )  \Big)  \\&\quad\quad\quad\quad\quad\quad\quad\quad\quad\quad\quad\quad\quad\quad\quad\quad\quad\quad\quad\quad\quad\quad\quad\text{for } k(i)\neq k(j) \\
        &\hspace{-0.3cm}0   \\&\quad\quad\quad\quad\quad\quad\quad\quad\quad\quad\quad\quad\quad\quad\quad\quad\quad\quad\quad\quad\quad\quad\quad\text{for } k(i)=k(j)
      \end{cases}\right\},\label{eq:comparison_matrix_def_K=N}
    \end{align}
for a fixed vector of hyperparameter $\xi$. The cardinality of the set $S(A)$ of ``strange'' points
\begin{align}
    \label{eq:def_strange_points}
    S(A) &= \bigg\{ i \in \{ 1,2,\ldots,N_\tau + 1 \} : \sum_{j'=1}^{N_\tau+1}A(i,j') \geq (1-\alpha')(N_\tau + 1)\bigg\}
\end{align}
can be bounded as $|S(A)| \leq N_\tau + 1 - (1-\alpha')(N_\tau +1)$ \cite{barber2021predictive, romano2020classification}. Therefore, theorem~\ref{ther:XB-CP} holds for $K=N_\tau$, since any $N_\tau+1$ points can be ``strange points'' with equal probability thanks to Assumption~\ref{assump:exchangeable}.

To address the case $K < N_\tau$, we follow \cite[Section B.2.2]{barber2021predictive} by drawing $N_\tau/K-1$ additional test examples that are all assigned to the  $(K+1)$th fold. This way, the actual $(N_\tau+1)$th test point is equally likely to be in any of the $K+1$ folds. Now, taking the augmented data set $\bar{\mathcal{D}}$ that contains all the $N_\tau+N_\tau/K$ examples in lieu of $\mathcal{D}$ in \eqref{eq:comparison_matrix_def_K=N}, we can bound the number of ``strange points'' in set \eqref{eq:def_strange_points} as
\begin{align}
    |S(A)| \leq N_\tau+N_\tau/K - (1-\alpha')(N_\tau + 1).
\end{align}
Finally, by using the same proof technique in \cite[Section B.2.2]{barber2021predictive}, we have the inequality
\begin{align}
    \label{eq:final_ineq_proof}
    \mathbb{P}\big( \mathbf{y}_\tau \in \Gamma_{\alpha}^{K\text{-XB}}(\mathbf{x}_\tau|\mathbfcal{D}_\tau,\xi)  \big) \geq  1 - \alpha' -\frac{1-K/N_\tau}{K+1}.
\end{align}
In Theorem~\ref{ther:XB-CP}, we choose $\alpha' = \frac{1-K/N_\tau}{K+1}$, which satisfies per-task validity condition (1) from \eqref{eq:final_ineq_proof}.

\subsection{Proof for Definition~\ref{def:soft_ineff}}
\label{app:proof_soft_ineff}
From the definition of the XB-CP set predictor \eqref{eq:cross_val_set_predictor_indicator}, the inefficiency can be obtained as
\begin{align}
\label{eq:precise_inefficiency}
     &|{\Gamma}_\alpha^{K\text{-XB}}(x_\tau|\mathcal{D}_\tau,\xi)| \nonumber\\&=  \sum_{y \in \mathcal{Y}_\tau} \mathbf{1} \bigg( \sum_{i=1}^{N_\tau} \mathbf{1} \Big( \min_{k\in\{ 1,\ldots,K\}}\text{NC}((x_\tau,y)|{\mathcal{D}}_{\tau,\neg k},\xi)  \leq  \text{NC}(z_\tau[i]|{\mathcal{D}}_{\tau, \neg k(i)},\xi)  \Big)\:\geq\: \lfloor \alpha'(N_\tau+1)\rfloor  \bigg) \nonumber\\
     &= \sum_{y \in \mathcal{Y}_\tau} \mathbf{1} \Big(  Q^-_{1-\alpha'}\Big(\Big\{\min_{k\in\{ 1,\ldots,K\}}\text{NC}((x_\tau,y)|{\mathcal{D}}_{\tau, \neg k},\xi)  -  \text{NC}(z_\tau[i]|{\mathcal{D}}_{\tau, \neg k(i)},\xi)    \Big\}_{i=1}^{N_\tau}\Big) \leq 0 \Big)  \\\label{eq:precise_inefficiency:2}
     &= \sum_{y \in \mathcal{Y}_\tau} \mathbf{1} \Big(  Q_{1-\alpha'}\Big(\Big\{ \text{NC}(z_\tau[i]|{\mathcal{D}}_{\tau, \neg k(i)},\xi) - \min_{k\in\{ 1,\ldots,K\}}\text{NC}((x_\tau,y)|{\mathcal{D}}_{\tau, \neg k},\xi)  \Big\}_{i=1}^{N_\tau}\Big) \geq 0 \Big),
\end{align}
with $Q_{1-\alpha}^{-}( \{ a[i] \}_{i=1}^M ):= -Q_{1-\alpha}( \{ -a[i] \}_{i=1}^M )$ being the $\lfloor \alpha(M+1)\rfloor$th smallest value in the set $\{a[1],\ldots,a[M],\infty\}$. The equality in \eqref{eq:precise_inefficiency} is proved as follows. Defining $g(z_\tau) := \min_{k\in\{ 1,\ldots,K\}}\text{NC}((x_\tau,y)|{\mathcal{D}}_{\tau,\neg k},\xi) $ with $z_\tau =(x_\tau,y)$ and   $ f(z_\tau[i]) := \text{NC}(z_\tau[i]|{\mathcal{D}}_{\tau, \neg k(i)},\xi) $,  we show that the inequality $\sum_{i=1}^{N_\tau}\mathbf{1}\big(g(z_\tau) \leq f(z_\tau[i]) \big) \geq \lfloor \alpha' (N_\tau+1)\rfloor$ is equivalent to $Q_{1-\alpha'}^{-}\big( \{ g(z_\tau)-f(z_\tau[i]) \}_{i=1}^{N_\tau} \big) \leq 0$. This is a consequence of the following equivalence relations:
\begin{align}
 &\sum_{i=1}^{N_\tau}\mathbf{1}\big(g(z_\tau) \leq f(z_\tau[i]) \big) \geq \lfloor \alpha' (N_\tau+1)\rfloor \nonumber\\
 \Leftrightarrow& \sum_{i=1}^{N_\tau}\mathbf{1}\big(g(z_\tau) - f(z_\tau[i]) \leq 0 \big) \geq \lfloor \alpha' (N_\tau+1)\rfloor \nonumber\\
 \Leftrightarrow& \text{ at least $\lfloor \alpha' (N_\tau+1)\rfloor$ values of $g(z_\tau) - f(z_\tau[i])$} \text{ are smaller than or equal to $0$ } \nonumber\\
 \Leftrightarrow& \text{$\lfloor \alpha' (N_\tau+1)\rfloor$th smallest value of $g(z_\tau)-f(z_\tau[i])$}\text{ is smaller than or equal to $0$} \nonumber\\
 \Leftrightarrow& Q_{1-\alpha'}^{-}\big( \{ g(z_\tau)-f(z_\tau[i]) \}_{i=1}^{N_\tau} \big) \leq 0
 \nonumber\\
 \Leftrightarrow& Q_{1-\alpha'}\big( \{ f(z_\tau[i])-g(z_\tau) \}_{i=1}^{N_\tau} \big) \geq 0.
\end{align}

By replacing $\mathbf{1}(\cdot)$ with the sigmoid $\sigma(\cdot)$, $\min(\cdot)$ with $\text{softmin}(\cdot)$ \eqref{eq:softmin}, and the quantile $Q_{1-\alpha'}(\cdot)$ with $\hat{Q}_{1-\alpha'}(\cdot)$ \eqref{eq:softquan}, we finally obtain the soft inefficiency of $K$-fold XB-CP predictor in \eqref{eq:soft_ineff_2} from \eqref{eq:precise_inefficiency:2}.

\section{Details on Soft Adaptive NC Scores}
\label{app:soft_adaptive_NC}
Recalling \eqref{eq:nc_adaptive_implement}, while denoting $p_{y'}:=p(y'|x,\tilde{\mathcal{D}}_\tau,\xi)$ and $p_{y}:= p(y|x,\tilde{\mathcal{D}}_\tau,\xi)$, the adaptive NC score for input-output pair $z=(x,y)$ with $x\in \mathcal{X}_\tau$ and $y \in \mathcal{Y}_\tau$ can be computed as
        \begin{align}  
            \label{eq:adaptive_preprocess}
            \text{NC}^\text{ada}(z|\tilde{\mathcal{D}}_\tau,\xi) &=\sum_{y'\in\mathcal{Y}_\tau} \mathbf{1} \big( p_{y'} \geq p_y  \big) p_{y'} \nonumber\\
            &=\sum_{y'\in\mathcal{Y}_\tau} \text{ReLU}(p_{y'} - p_y)  + p_y\sum_{y' \in \mathcal{Y}_\tau} \mathbf{1}( p_{y'} \geq p_y) \nonumber \\
            &=1+\sum_{y'\in\mathcal{Y}_\tau} \text{ReLU}(p_{y} - p_{y'})  - p_y\sum_{y' \in \mathcal{Y}_\tau} \mathbf{1}( p_{y'} < p_y).
        \end{align}
    We define the soft adaptive NC score by approximating the indicator function $\mathbf{1}(\cdot)$ with the sigmoid $\sigma(\cdot)$ as
    \begin{align}    
            \text{NC}^\text{ada}(z|\tilde{\mathcal{D}}_\tau,\xi) =1+\sum_{y'\in\mathcal{Y}_\tau} \text{ReLU}(p_{y} - p_{y'})  - p_y\sum_{y' \in \mathcal{Y}_\tau} \sigma( p_{y'} < p_y).
        \end{align}
    Note that, we have found that the preprocessing \eqref{eq:adaptive_preprocess} yields better empirical per-input coverage as compared to the direct approximation of \eqref{eq:nc_adaptive_implement} that replaces the indicator function $\mathbf{1}(\cdot)$ with sigmoid $\sigma(\cdot)$, i.e., $\sum_{y'\in\mathcal{Y}_\tau} \sigma \big( p_{y'} \geq p_y  \big) p_{y'}$.

\section{Additional Experiments}
\subsection{Demodulation}
\label{app:demodulation}
\begin{figure}[t]
 \begin{center}\includegraphics[scale=0.16]{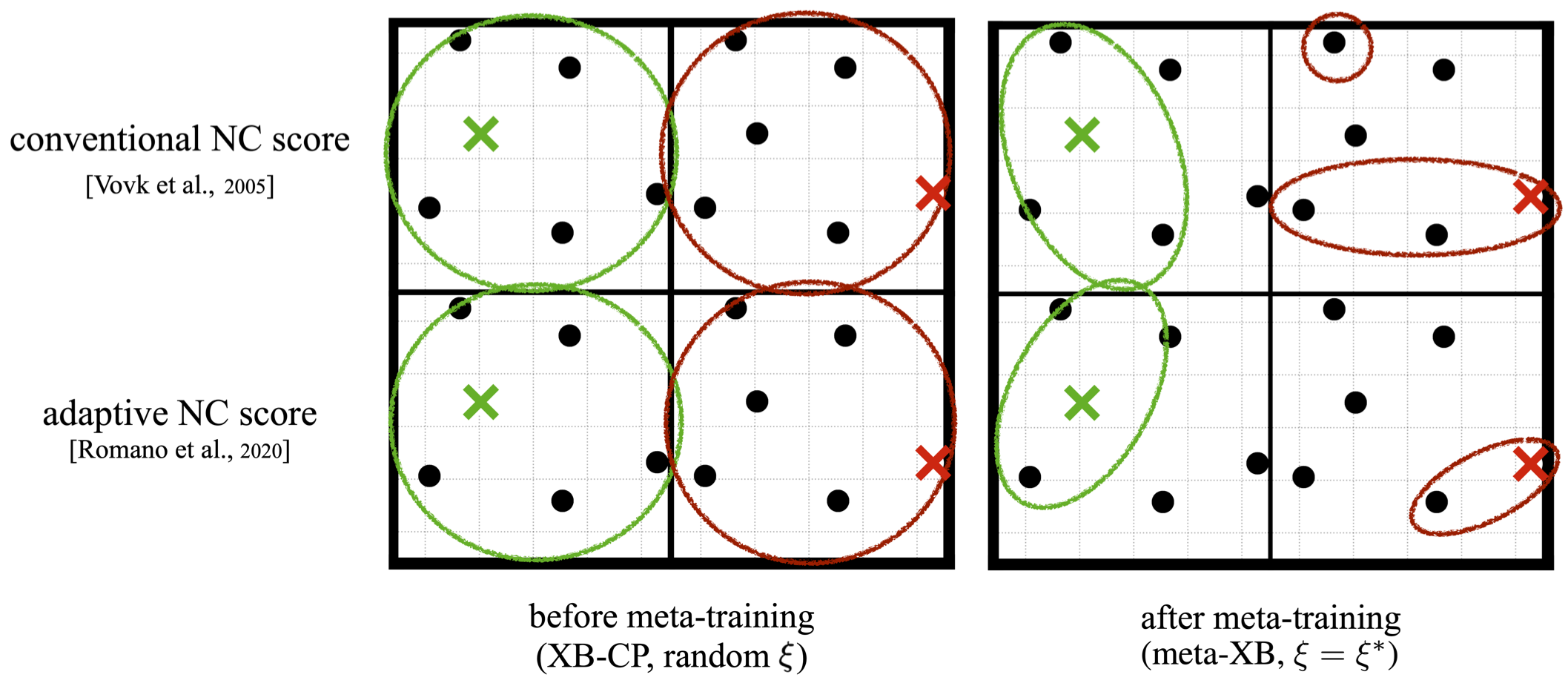} 
   \caption{Illustration of set predictions for the demodulation problem described in Appendix~\ref{app:demodulation}. Colored crosses represent ground-truth outputs and the correspondingly colored circles depict the predicted sets. For visualization purpose, the figure is generated based on a single realization of task $\tau$, data set $\mathcal{D}_\tau$, and test input $x_\tau$. $T=1000$ different tasks are used for meta-training.} \label{fig:gam}
   \end{center}
 \vspace{-0.35cm}
\end{figure}
To elaborate further on the last column in Fig.~\ref{fig:cartoon}, here we present a toy example that allows us to visualize the set predictors obtained by XB-CP and meta-XB, both with conventional and adaptive NC. To this end, we implement XB-CP with the same neural network used for meta-XB but with the hyperparameter vector $\xi$ defining the initialization of GD set to a random vector \cite{he2015delving}. Given a learning task $\tau$, the input and output space $\mathcal{X}_\tau$ and $\mathcal{Y}_\tau$ are given by the set of complex points, i.e., by the two-dimensional real vectors \cite{larsson2017golden}
\begin{align}
    &\mathcal{X}_\tau = \mathcal{Y}_\tau =\left\{ \sqrt{\frac{2z}{M+1}} e^{j 2\pi \big(1-\frac{\sqrt{5}-1}{2}\big) z} e^{j \phi_\tau} \text{ for } z=1,2,\ldots,M\right\},
\end{align}
for some task-specific phase shift $\phi_\tau \in [0,2\pi]$. Denoting as $\mathcal{N}_\tau(x) = \{ y \in \mathcal{Y}_\tau: |x-y| \leq r \text{ and } y \neq x \} $ the set of neighboring points within some radius $r$, the ground-truth distribution $p(y_\tau|x_\tau)$ is such that $y_\tau$ equals $x_\tau$ with probability $1-p$, and it equals any neighboring point $y \in \mathcal{N}_\tau(x)$ with probability $p/|\mathcal{N}_\tau(x)|$.  We set $M=6, r=1.3,$ and $p=0.2$. We design neural network classifier to consist of two hidden layers with Exponential Linear Unit (ELU) activation \cite{clevert2015fast} in the hidden
layers and a softmax activation in the last layer.

Fig.~\ref{fig:gam} visualizes the set predictors for XB-CP, i.e., with a random hyperparameter vector $\xi$, and for meta-XB, after meta-training with $1000$ tasks, by focusing on a specific realizations of phase shift that follows the distribution $\phi_\tau \sim \text{Unif}[0,2\pi)$. By transferring knowledge from multiple tasks, meta-XB  is seen to yield more efficient set predictors. Furthermore, by using adaptive NC scores, meta-XB can adjust the prediction set size depending on the ``difficulty'' of classifying the given input, while a conventional NC score tends to produce set predictors of similar sizes across all inputs. Note, in fact, that inputs close to the center of the set, as the green example in Fig.~\ref{fig:gam}, have more neighbors as compared to points at the edge, as the red points in Fig.~\ref{fig:gam}, making it harder to identify the true value of $y$ given $x$.

\bibliographystyle{IEEEtran}
\bibliography{ref.bib}

\ifCLASSOPTIONcaptionsoff
  \newpage
\fi

\end{document}